\documentclass{article}

\PassOptionsToPackage{numbers, compress}{natbib}

\usepackage[preprint]{neurips_2023}
\usepackage{multirow}

\usepackage[utf8]{inputenc} 
\usepackage[T1]{fontenc}    
\usepackage{hyperref}       
\usepackage{url}            
\usepackage{booktabs}       
\usepackage{amsfonts}       
\usepackage{nicefrac}       
\usepackage{microtype}      
\usepackage{xcolor}         
\usepackage{graphicx}

\title{Understanding the Benefits of Image Augmentations}

\author{%
  Matthew Iceland\\
  University of Rochester\\
  \texttt{miceland@u.rochester.edu} \\
   \And
   Christopher Kanan \\
   University of Rochester \\
   \texttt{ckanan@cs.rochester.edu} \\
}

\begin{document}

\maketitle

\begin{abstract}
  Image Augmentations are widely used to reduce overfitting in neural networks. However, the explainability of their benefits largely remains a mystery. We study which layers of residual neural networks (ResNets) are most affected by augmentations using Centered Kernel Alignment (CKA). We do so by analyzing models of varying widths and depths, as well as whether their weights are initialized randomly or through transfer learning. We find that the pattern of how the layers are affected depends on the model's depth, and that networks trained with augmentation that use information from two images affect the learned weights significantly more than augmentations that operate on a single image. Deeper layers of ResNets initialized with ImageNet-1K weights and fine-tuned receive more impact from the augmentations than early layers. Understanding the effects of image augmentations on CNNs will have a variety of applications, such as determining how far back one needs to fine-tune a network and which layers should be frozen when implementing layer freezing algorithms.
\end{abstract}

\section{Introduction}

Deep learning is an effective method for computer vision tasks such as image classification and object detection. One of the greatest challenges in computer vision is overfitting, as models may easily achieve perfect or near-perfect performance on a training set but fail on images it hasn't seen before. Several regularization methods exist to improve the generalization accuracy of deep neural networks, such as weight decay, dropout, transfer learning, and one-shot and zero-shot learning. Data augmentations reduce overfitting by increasing the size of the training set. Data warping performs transformations to the images while preserving their original labels, while oversampling creates synthetic images and adds them to the training set \cite{10.1007/s10462-021-10066-4}. Example of warping include geometric and color-space transformations, and examples of oversampling are mixup and generative adversarial networks (GANs).

Image augmentations have gained increasing attention be researchers over time. From 2015 to 2020, the number of research papers about image augmentations increased by a factor of almost 24, from 52 to 1269 \cite{10.1007/s10462-021-10066-4}. In particular, there is high interest in data augmentations within the medical field, as deep networks rely on large amounts of medical images for diagnosis, which can be difficult to obtain. \cite{10.1007/978-3-319-24574-4_28} and \cite{7426413} were considered breakthroughs in medical imaging, which demonstrated the effectiveness of convolutional neural networks (CNNs) that rely extensively on image augmentations for image segmentation. The COVID-19 pandemic in particular highlighted the issue of limited datasets when trying to identify the virus in medical images \cite{10.1007/s10462-021-10066-4}.

At a high level, image augmentations are thought to help networks "imagine" alternations to images and hence better understand them \cite{article}. More specifically, image augmentations are designed to make models invariant to factors such as viewpoint, lighting, and background \cite{10.1007/s10462-021-10066-4}. Augmentations are ubiquitously used in deep learning, yet it is unknown precisely how they affect the layers of a network during training. This study aims to elucidate which hidden layers of CNNs, or, more specifically, residual neural networks (ResNets) \cite{he2015deep}, are most affected by data augmentations on the image space. We rely on the use of centered kernel alignment (CKA) \cite{kornblith2019similarity} to evaluate the similarity between representations learned by ResNets to ultimately determine which layers receive the greatest benefit.

The knowledge of which layers are most effective should also be of value to the development of strong continual learning models that freeze the weights of some of its hidden layers when training on new data, such as REMIND \cite{https://doi.org/10.48550/arxiv.1910.02509} and SIESTA \cite{harun2023siesta} . The layers that receive the most benefit from image augmentations should ideally remain plastic during training. Additionally, by measuring the impact of augmentations when fine-tuning, we can discern how far back a model needs to be fine-tuned. For instance, if the early layers tend to be hardly affected by image augmentations, they could be omitted during training, thus saving compute.

This paper makes the following contributions:

\begin{enumerate}
\item {Deep layers tend to be more affected by data augmentations than early layers when fine-tuning networks.}
\item {Augmentations involving the combination of two images (i.e. mixup and cutmix) impact neural networks considerably more than augmentations transforming only a single image.}
\item {The number of layers in a convolutional neural network plays a significant role in how they are affected by augmentations.}
\end{enumerate}



\section{Related works}
\label{gen_inst}

\cite{tang2022explaining} studied how image augmentations, particularly, how random cropping, gaussian blurring, random scaling, distortion, and random erasing affect the salience maps of a small CNN trained on a small fraction of the MNIST dataset. They found that these basic image augmentations could significantly alter the saliencey mapping and the amount of information gained and lost from the network using Pearson's Correlation Coefficient (CC) and Kullback Leibler Divergence (KL). Furthermore, no correlation was found to exist between CC/KL and test accuracy.

Using a comparison metric based on how two networks use training data, \cite{shah2022modeldiff} compares two ResNet-18s trained with and without image augmentations on the LIVING17 \cite{santurkar2020breeds} dataset and concludes that standard augmentations (i.e. horizontal flips and random crops) can amplify texture biases and specific cases of co-occurance biases. Hence, data augmentations alter the relative importance of learned features \cite{shah2022modeldiff}.

Performing image augmentations such as random rotations and flips are thought to increase the diversity of limited datasets \cite{PANWAR2020109944}, which should make the training distribution more closely match the true, real-world data distribution. Nevertheless, the impact of augmentations on generalization capabilities largely remains a black box. In fact, \cite{elgendi2021effectiveness} revealed that the removal of geometric augmentations from several models actually increased their Matthew's correlation coefficient (MCC). Also, most papers studying augmentations focus on their resulting performance gains. We hope to better understand how these augmentations affect the network through layer-by-layer analysis and how this impact varies accross different model architectures.

Knowledge of the roles of each layer in the network can be highly valuable when designing image augmentations. For example, it is widely understood that later layers in CNNs tend to capture high-level features of images while earlier layers capture the low-level features. Such information is highly useful for tasks such as malignant melanoma detection. In order to generate more data for malignant skin lesions, the preceding knowledge can be used to design a style transfer algorithm where one of the first layers of a generative CNN may be responsible for generating an image style and one of the last layers responsible for its content \cite{8388338}. Doing so will allow sufficient data generation for malignant lesions, which are classified using both its shape (content) and color (style).

\section{Method}
\label{headings}

\begin{figure}
    \includegraphics[width=.33\textwidth]{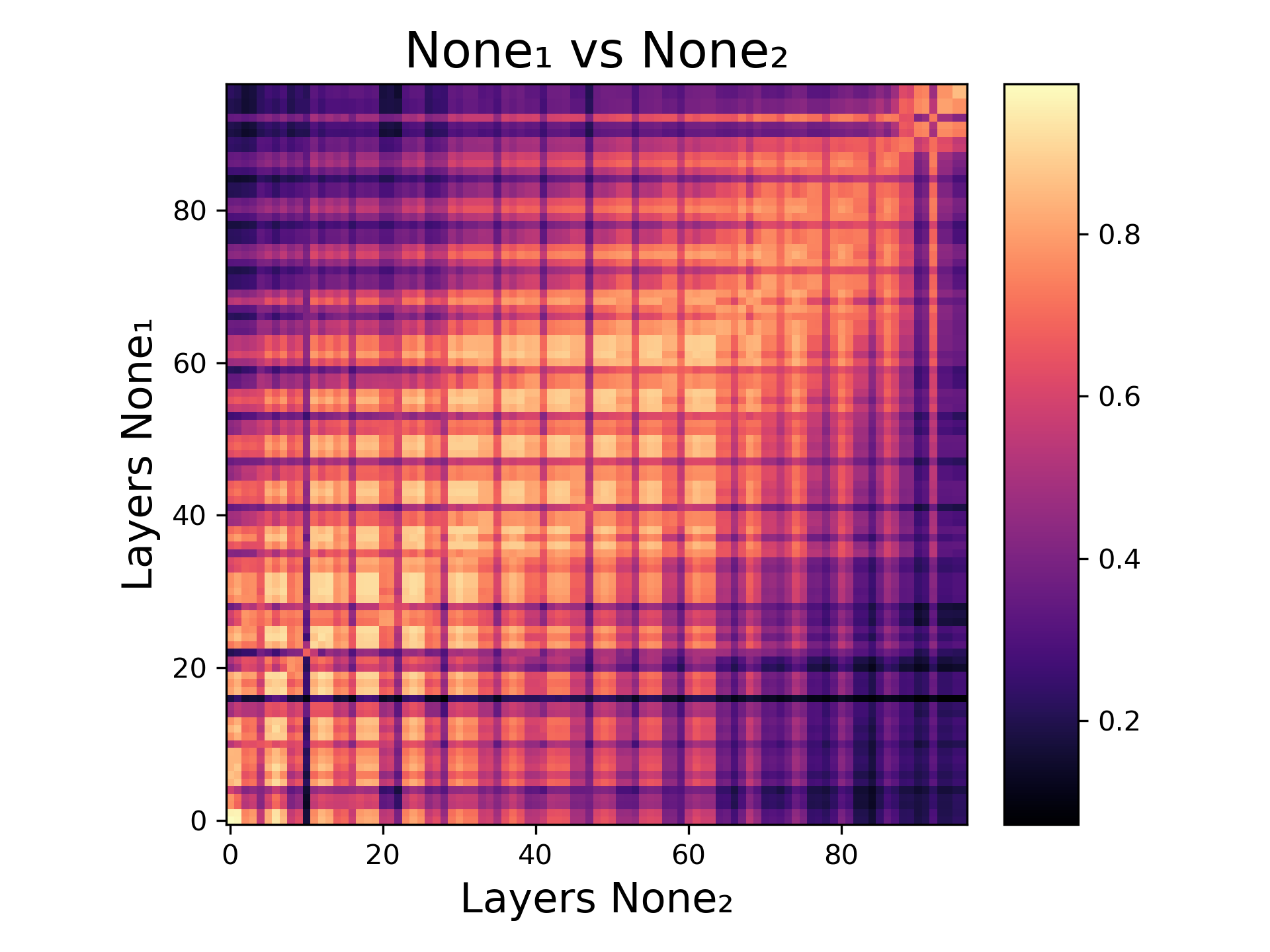}\hfill
    \includegraphics[width=.33\textwidth]{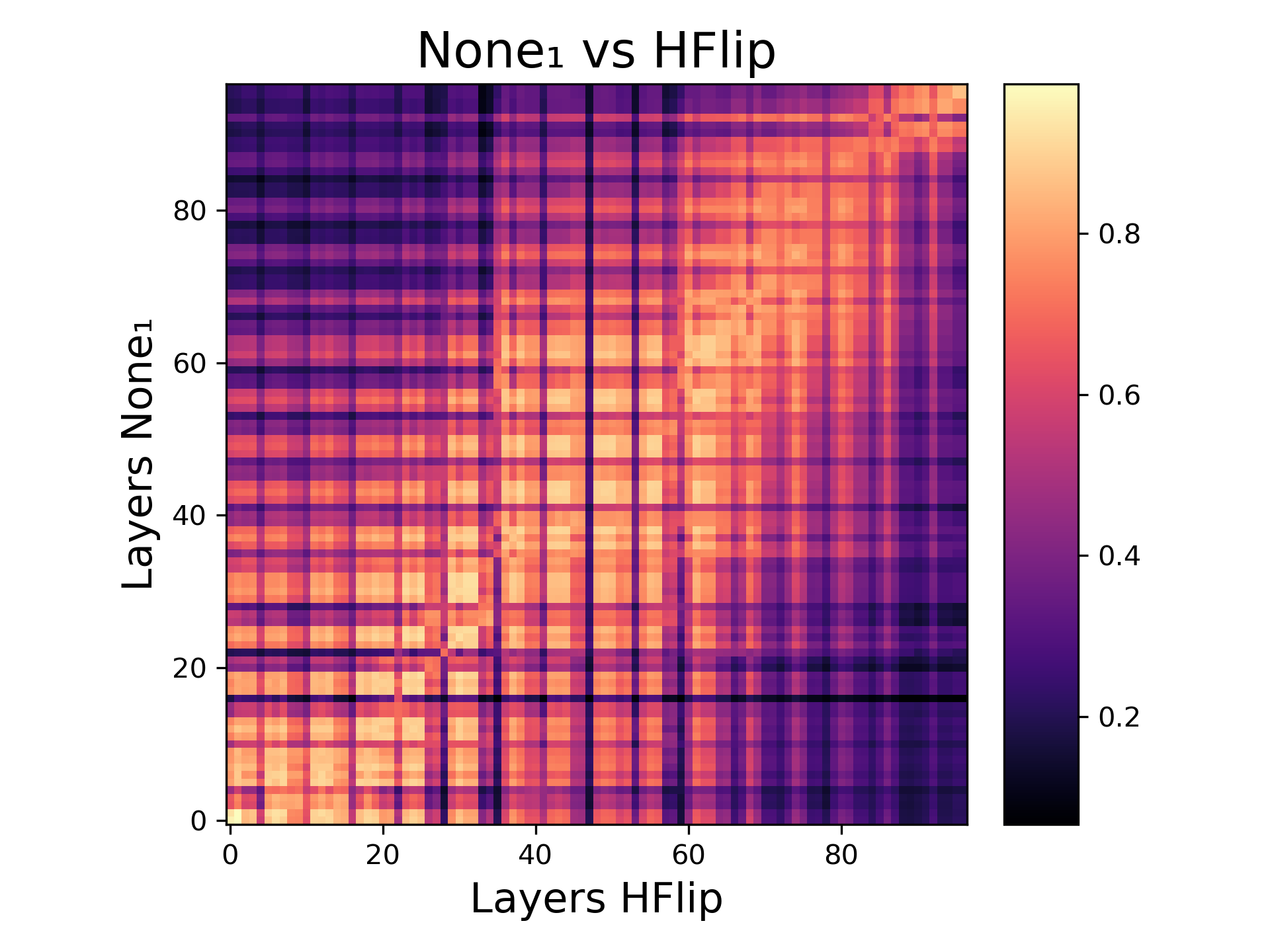}\hfill
    \includegraphics[width=.33\textwidth]{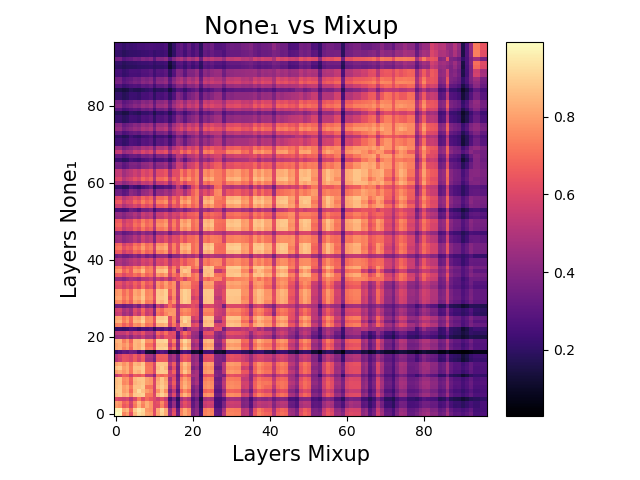}\hfill
    \caption{Matrices comparing the representations learned between all pairs of layers of two ResNet-32 models train without augmentations (left), one ResNet-32 trained with horizontal flips and another without augmentations (center), and a ResNet-32 trained with mixup and one without augmentations (right).}\label{fig:foobar}
\end{figure}

\subsection {Measuring impact from augmentations}

In order to measure which convolutions layers are most affected by augmentations, we train each network architecture (e.g. ResNet-18) twice with no augmentations but with different initializations, and then ten more time, each time with a different augmentation or combination of augmentations. The CKA is computed between each of the respective convolutional layers of the two networks trained without augmentations, hereafter referred to as \(none_{1}\) and \(none_{2}\). We train ResNet-32 and ResNet-56, which only have 0.46 million and 0.85 million parameters, respectively, from scratch on CIFAR-10. ResNet-18 and ResNet-50, which have about 11 million and 23 million parameters, are initialized with IMAGENET-1K weights and fine-tuned on Tiny-Imagenet-200.

Centered Kernel Alignment (CKA) \cite{kornblith2019similarity} is a measurement of of the similarity of two vectors or matrices of arbitrary dimensions. Given two embeddings, a CKA index from zero to one is calculated between them, with a score of zero indicating complete independence, and a score of one indicating identical matrices. CKA calculates the normalized Hilbert-Schmidt Independence Criterion (HSIC) between two sets of variables \cite{kornblith2019similarity}, which is a biased estimator of their independence. Furthermore, CKA can be calculated over a minibatch with its output index being independent of the batch size \cite{nguyen2021wide}.

Each of the ten augmented networks get compared with \(none_{1}\) and then \(none_{2}\). The amount of impact each layer receives from a particular augmentation is measured as the percent decrease in the CKA index from the comparison of \(none_{1}\) and \(none_{2}\) to the average similarity between a network trained with the augmentation and each network trained with no augmentations. Specifically, the impact of a given network trained with an augmentation \(aug\) is given by the following expression:

\[
\frac{CKA(none_{1}, none_{2}) - \frac{1}{2}(CKA(none_{1}, aug) + CKA(none_{2}, aug))} {CKA(none_{1}, none_{2})}
\]

\subsection {Image augmentations studied}

A variety of augmentations on the image space will be studied. Horizontal flips, applied with probability 0.5 for each image, and random crops are ubiquitously used for CNNs. The effects of these augmentations will be studied individually as well as in combination. For CIFAR-10, a random crop of size 32 and padding of 4 will be used on all training images, while a random resized cop will be used for Tiny-Imagenet-200 images with probability 0.5. These are implemented using the torchvision RandomHorizontalFlip(), RandomCrop(), and RandomResizedCrop() built-in methods.

Three color space augmentations wil also be studied. These include brightness jittering with a factor of 0.5, hue jittering with a factor of 0.15, and solarizing with a pixel value threshold of 127 and probability 0.5 of being applied. The brightness, hue, and solarize augmentations are implemented using the torchvision ColorJitter() and RandomSolarize() methods.

The models will also be trained using cutout, where a random 16 x 16 square of the image will be replaced with a gray square box. Although small, these cutouts can produce a noticable increase in performance. Mixup is another effective augmentation that combines the pixel values of two random images, and the ground truth label of the new image is a linear interpolation of the original one-hot labels. The combination ratio of the two images is sampled from a beta distribution paramaterized by a variable alpha. Cutmix is similar to mixup, except that only a random box of the final image will be mixed. Both mixup and cutmix are applied with an alpha value of 1.0 in this study. As implemented in the original cutmix paper \cite{yun2019cutmix}, the cutmix transformation is applied with probability 0.5 for CIFAR-10 and probability 1.0 for ImageNet or Tiny-Imagenet-200.

Lastly, AutoAugment \cite{cubuk2019autoaugment} is studied in this paper. Given a policy as a parameter (e.g. a policy for CIFAR or ImageNet), AutoAugment randomly chooses a subpolicy for each minibatch. A subpolicy consists of two image augmentations, such as translations, rotations, and shears. AutoAugmentation policies may improve network generalization by percentage points that rival other effective augmentation combinations, such as horizontal flip and random crop.

\section{Setup}

\begin{figure}
    \includegraphics[width=.40\textwidth]{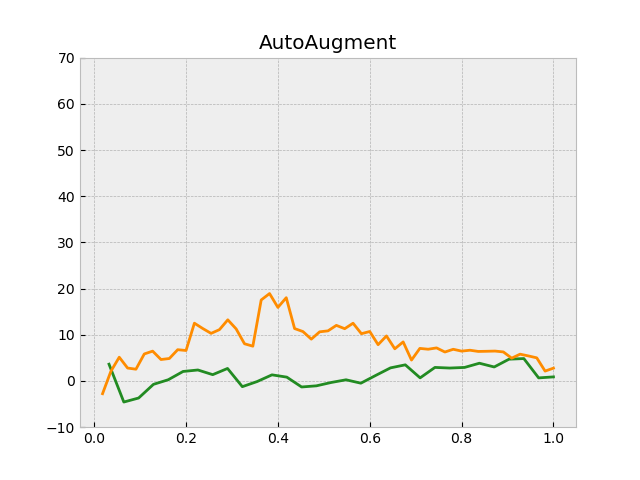}\hfill
    \includegraphics[width=.40\textwidth]{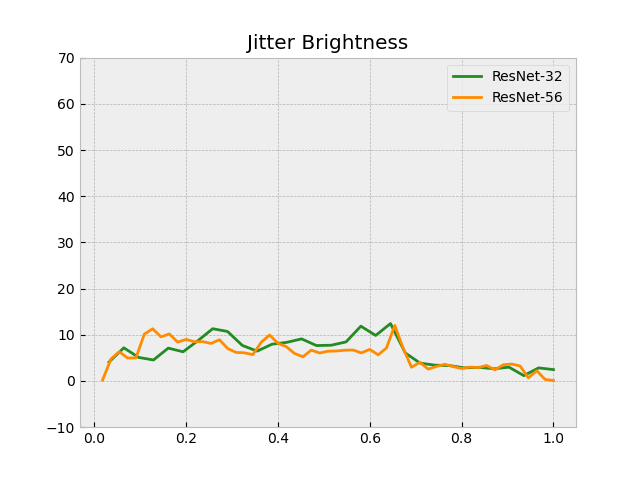}
    \\[\smallskipamount]
    \includegraphics[width=.40\textwidth]{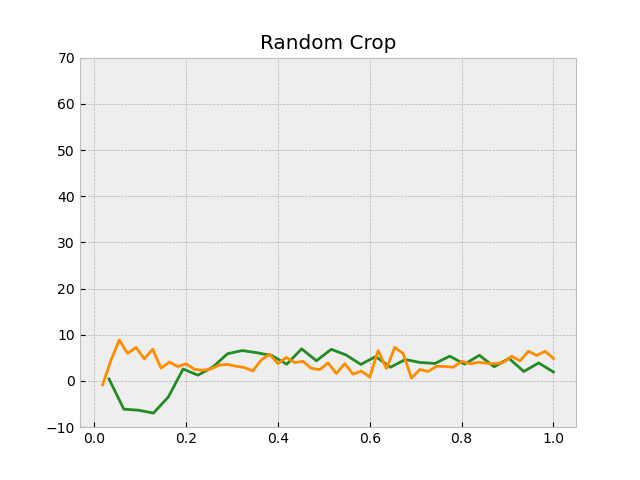}\hfill
    \includegraphics[width=.40\textwidth]{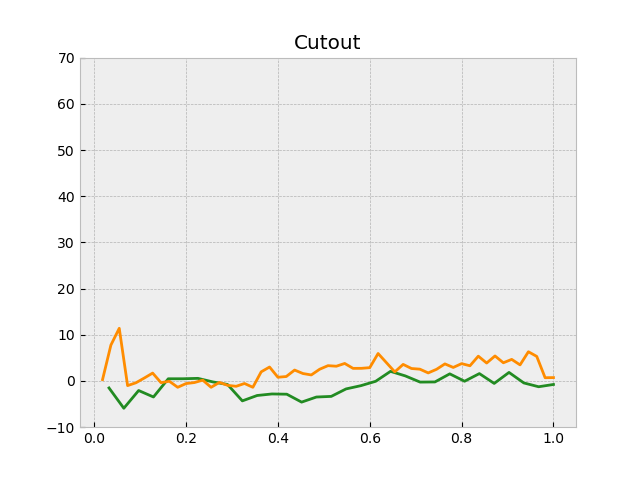}
    \\[\smallskipamount]
    \includegraphics[width=.40\textwidth]{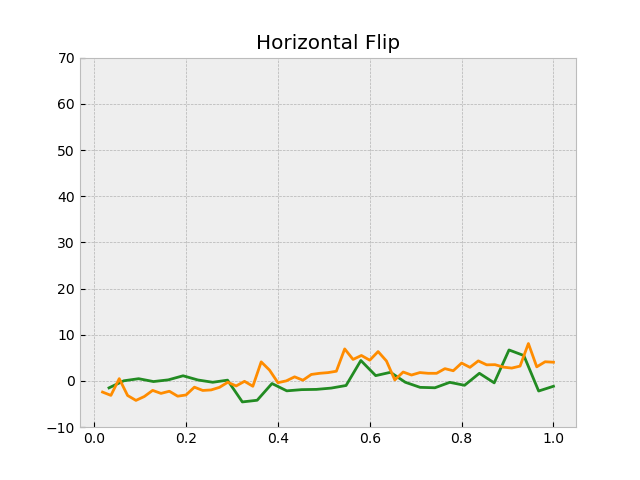}\hfill
    \includegraphics[width=.40\textwidth]{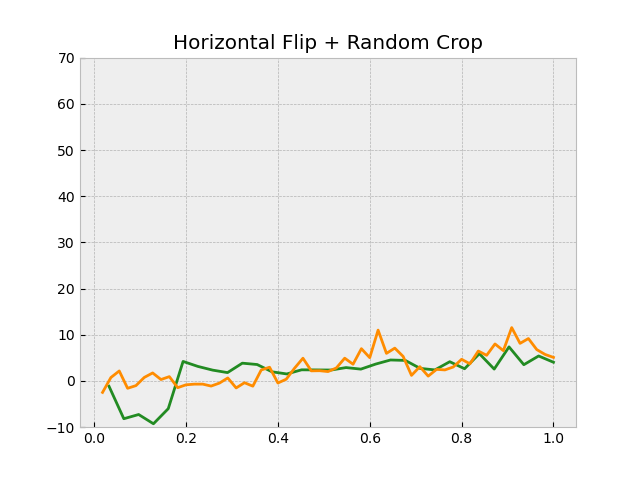}
    \\[\smallskipamount]
    \includegraphics[width=.40\textwidth]{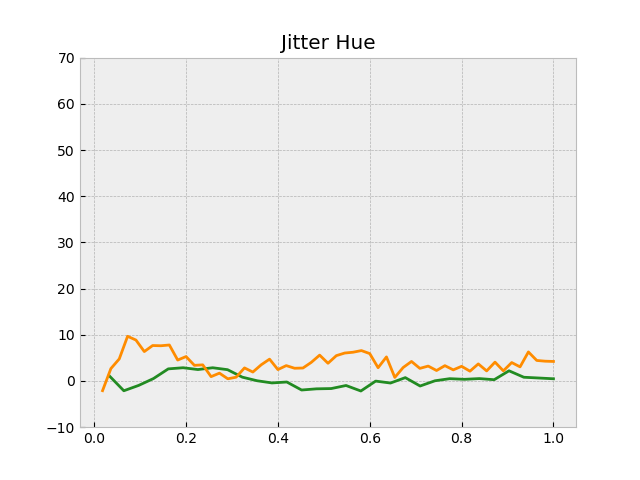}\hfill
    \includegraphics[width=.40\textwidth]{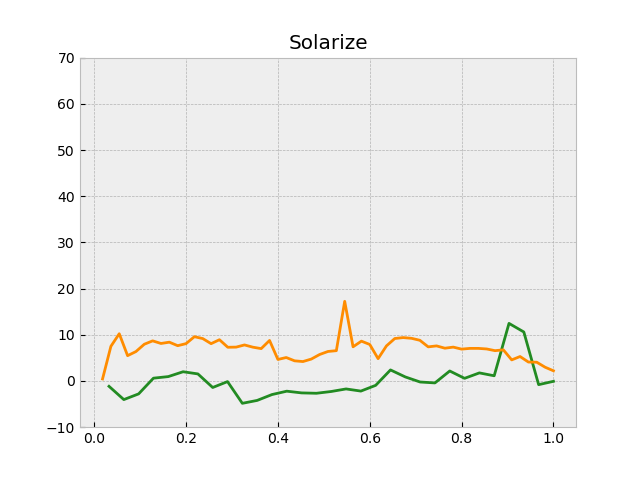}
    \\[\smallskipamount]
    \includegraphics[width=.40\textwidth]{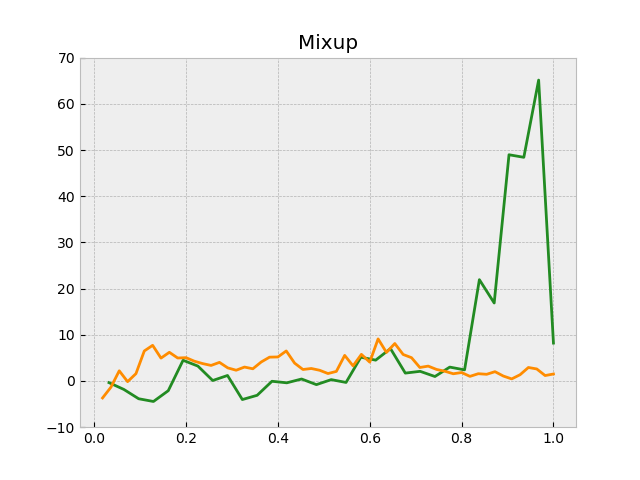}\hfill
    \includegraphics[width=.40\textwidth]{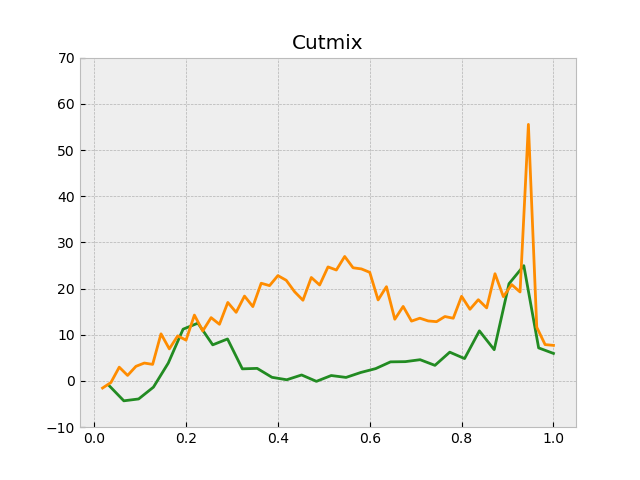}
    \caption{Percent decrease in CKA similarity for each convolutional layer of ResNet-32 (green) and ResNet-56 (orange). The horizontal axis is the normalized depth of each layer from zero to one.}
\end{figure}

ResNet-32 and ResNet-56 are trained on CIFAR10 while ResNet-18 and ResNet-50 are trained on Tiny-Imagenet-200. Each model is trained with the stochastic gradient descent (SGD) optimizer with a weight decay factor of 1e-4 and momentum of 0.9. The models are trained for 200 epochs, and the model is rolled back to whichever state achieved the greatest validation accuracy. The models fine-tuned on Tiny-Imagenet-200 are trained with a learning rate of 1e-3, while those trained on CIFAR-10 use a learning rate of 1e-1. 

For each model being considered, it is trained a total of 12 times on its respective dataset: twice to compute the none1 and none2 networks and ten times for each of the augmentations or combination of augmentations. After all models are fully trained for a given architecture, each is compared with its respective \(none_{1}\) and \(none_{2}\) models, and the CKA between none1 and none2 is also calculated, resulting in matrices such as those visualized in Figure 1. For each architecture, the CKA is computed as described in Section 3 for each convolutional layer across all the augmented networks. For each model studied, the corresponding validation set of its training dataset is used to produce the embeddings to calculate CKA.

\section{Results}

\begin{figure}
    \includegraphics[width=.40\textwidth]{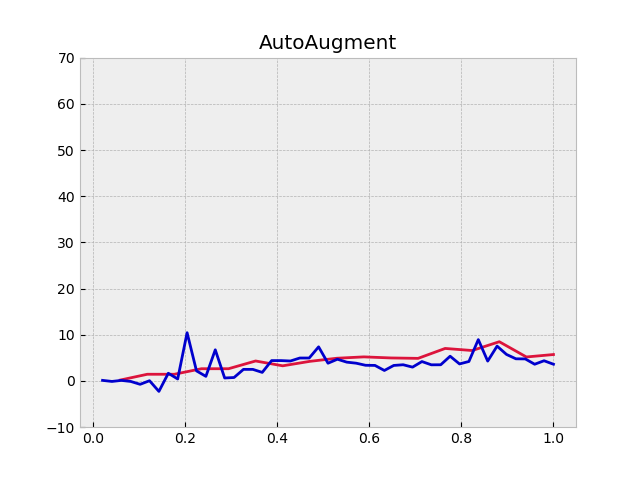}\hfill
    \includegraphics[width=.40\textwidth]{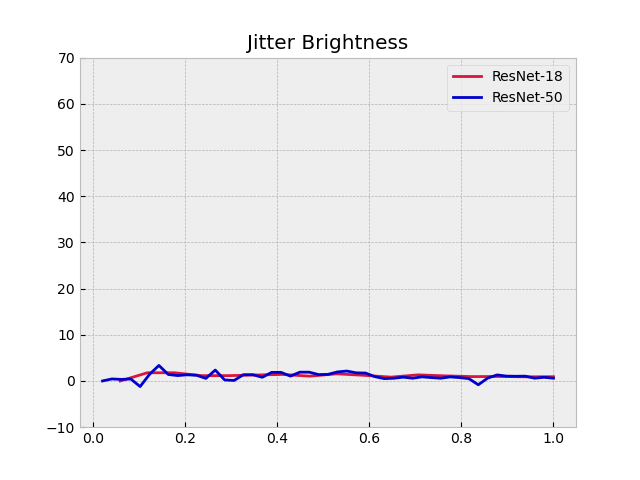}
    \\[\smallskipamount]
    \includegraphics[width=.40\textwidth]{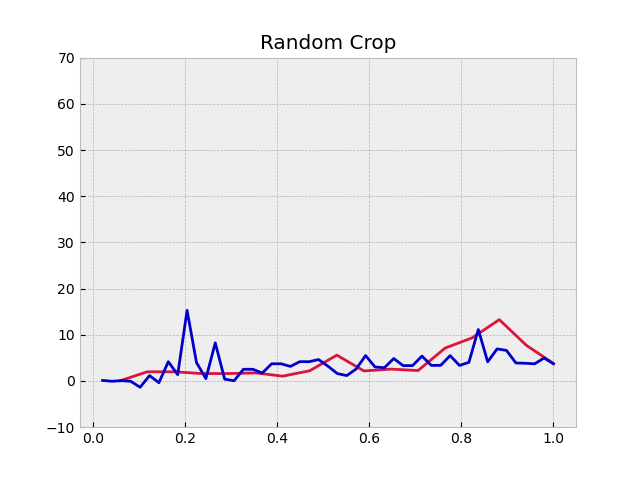}\hfill
    \includegraphics[width=.40\textwidth]{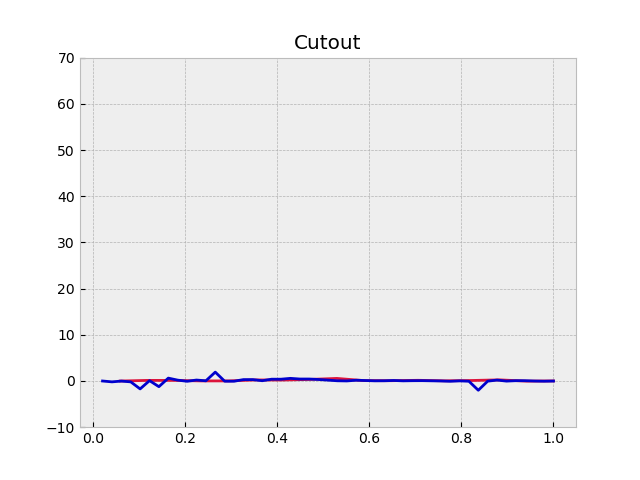}
    \\[\smallskipamount]
    \includegraphics[width=.40\textwidth]{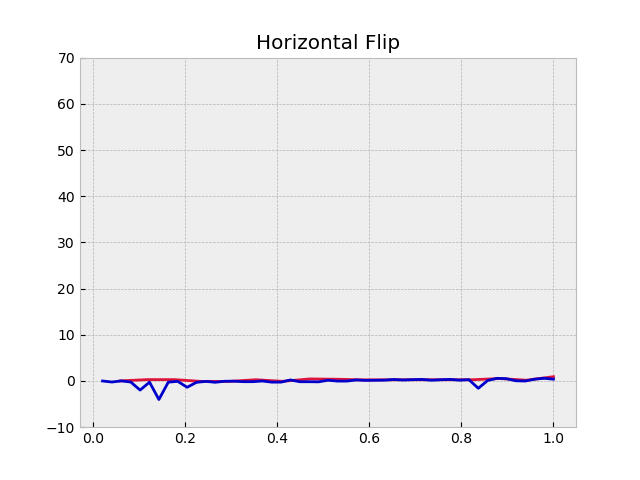}\hfill
    \includegraphics[width=.40\textwidth]{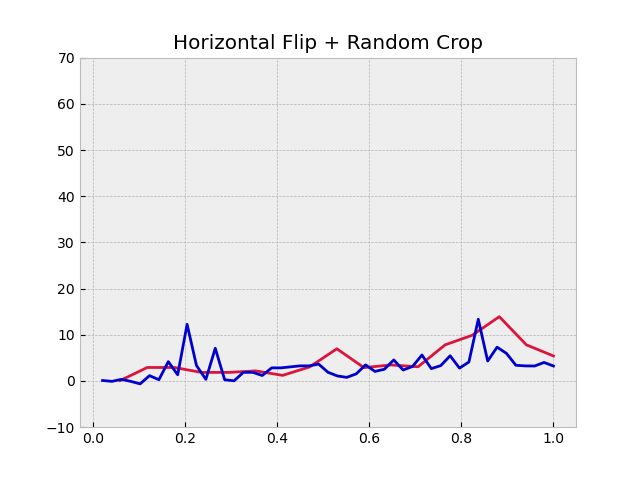}
    \\[\smallskipamount]
    \includegraphics[width=.40\textwidth]{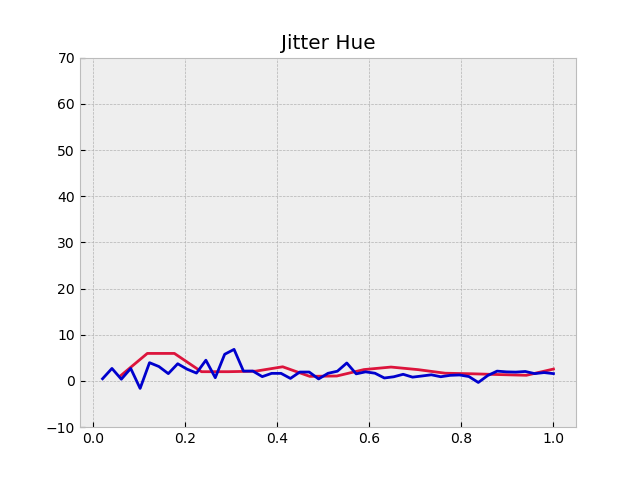}\hfill
    \includegraphics[width=.40\textwidth]{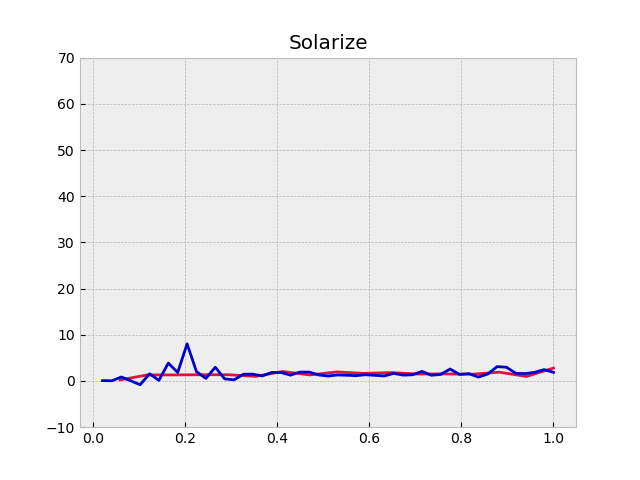}
    \\[\smallskipamount]
    \includegraphics[width=.40\textwidth]{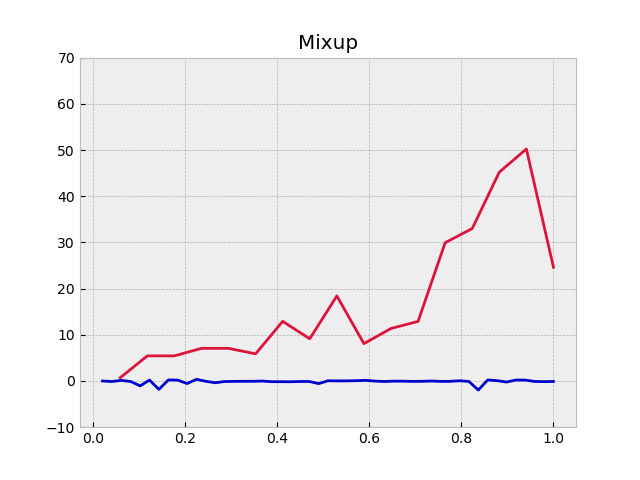}\hfill
    \includegraphics[width=.40\textwidth]{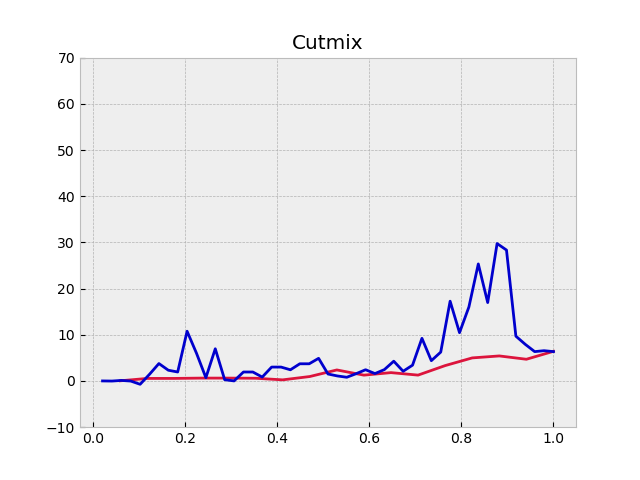}
    \caption{Percent decrease in CKA similarity for each convolutional layer of ResNet-18 (red) and ResNet-50 (blue). The horizontal axis is the normalized depth of each layer from zero to one.}
\end{figure}

\subsection{Main findings}

\begin{table}
  \centering
  \caption{Test accuracy and average percent decrease in similarity for each model and augmentation}
  \begin{tabular}{l|l|l|l|l|l|l|l|l}
    \hline
    \multirow{2}{*}{Augmentation} &
      \multicolumn{2}{c}{ResNet-32} &
      \multicolumn{2}{c}{ResNet-56} &
      \multicolumn{2}{c}{ResNet-18} &
      \multicolumn{2}{c}{ResNet-50} \\
    & Acc. & Dec. & Acc. & Dec. & Acc. & Dec. & Acc. & Dec.\\
    \hline
    AutoAugment & 88.83 & 1.17 & 88.63 & 8.13 & 71.01 & 4.32 & 77.01 & 3.39\\
    \hline
    Jitter Brightness & 82.37 & 6.37 & 83.20 & 5.74 & 71.24 & 1.15 & 77.47 & 1.02\\
    \hline
    Random Crop & 90.64 & 2.93 & 89.51 & 3.89 & 72.62 & 3.89 & 78.73 & 3.46 \\
    \hline
    Cutout & 87.84 & -1.12 & 87.76 & 2.26 & 71.68 & 0.14 & 77.64 & 0.04\\
    \hline
    Horizontal Flip & 87.39 & -0.12 & 85.94 & 1.25 & 72.90 & 0.26 & 78.38 & -0.12\\
    \hline
    HFlip + RCrop & 92.02 & 1.83 & 91.01 & 2.94 & 73.18 & 4.56 & 79.22 & 3.05 \\
    \hline
    Jitter Hue & 84.97 & 0.28 & 80.58 & 3.93 & 70.13 & 2.40 & 76.48 & 1.84\\
    \hline
    Solarize & 83.83 & -0.01 & 83.80 & 7.02 & 70.53 & 1.51 & 76.87 & 1.57\\
    \hline
    Mixup & 88.19 & 7.26 & 80.78 & 3.28 & 49.75 & 16.91 & 77.40 & -0.14 \\
    \hline
    Cutmix & 88.38 & 4.92 & 88.52 & 15.63 & 71.27 & 2.11 & 77.97 & 5.75-\\
    \hline
    
  \end{tabular}
\end{table}

The main results of this paper can be summarized by the graphs of Figure 1 and Figure 2. The impact of each augmentation on each layer of a network corresponds to the heights of the points on the graphs. Impact can also be thought of as a decrease in CKA, which measures the amount of change each layer received from the augmentation compared to the baseline (i.e. the CKA between \(none_{1}\) and \(none_{2}\)). A positive decrease in CKA corresponds to the augmented networks being more different to the none networks the the none networks are to each other. A negative decrease indicates that the augmented networks are, on average, more similar to the none networks than the none networks are to each other.

From Figures 2 and 3, mixup and cutmix clearly affect the convolutional layers more than the other augmentation. From Table 1, the greatest average percent decreases in CKA similarity are 15.63 for cutmix on ResNet-56 and 16.91 for mixup on ResNet-18. Particularly, there tend to be large spikes in decrease in similarity at layers towards the end of the network. Generally, the deeper layers of the fine-tuned ResNet-18 and ResNet-50 tend to receive more benefit than the early layers, as is apparant from the graphs of AutoAugment, random crop, horizontal flip plus random crop, mixup, and cutmix in Figure 3. However, the pattern does not hold consistently for the smaller residual networks trained on CIFAR-10. Augmentations such as AutoAugment, cutmix, and solarizing considerably affect the first half of the layers. Overall, the networks trained from scratch see changes in both early and deep layers from augmentations.

Also, the depth and number of parameters in the model may significantly alter how it learnes from the augmentations. For instance, ResNet-18 sees substantial benefits from mixup while ResNet-50 receives little, and the reverse is true for cutmix. Further, seemingly disparate image transformations, such as geometric versus color-space augmentations, don't necesarily affect CNNs with disparate patterns. For example, the graphs for cutout, horizontal flip, and jittering hue in Figure 2 look relatively similar.

\subsection{The connection between CKA and performance}

The connection between a network's top-1 generalization accuracy and the average impact each of its layers receive also must be examined. We find that higher validation accuracy doesn't necessarily correspond to a greater percent deacrease in CKA over the layers. As can be seen from Table 1, ResNet-50 trained with horizontal flips achieves the second-highest accuracy for ResNet-50 but sees the second-lowest decrease in similarity. Also, ResNet-32 trained with horizontal flips and random crops achieves significantly greater performance than jittering brightness, yet the former sees noticeably less impact.

Nevertheless, for the particular augmentation of random cropping, there appears to be a connection between accuracy and CKA change, for it achieves some of the highest accuracies on ResNets 18 and 50, both alone and with horizontal flip, and these correspond to some of the greatest percent similarity decreases. However, horizontal flips, surprisingly, produce some of the highest accuracies for the larger ResNets but appear to affect the networks very minutely. Given how certain augmentations may actually harm the performance of deep neural networks, it is not surprising that there doesn't appear to be a strong correspondence between accuracy and impact.

\section{Discussion}

One would naturally wonder how significant these differences in impact are. Does a layer that receives a decrease in CKA of 50\% learn significantly more different representations than it would without augmentations compared to a layer with a CKA decrease of only 5\%?

Feature maps can be used to visualize such differences. Take, for example, the 20th convolutional layer of ResNet-56, which sees a 21.16\% drop in similarity for cutmix yet only a 1.99\% drop for cutout. In Figure 4, the feature maps for a sample image of a hen are displayed for this layer. Though cutmix affects this layer much more than cutout, the feature map of cutmix doesn't appear significantly more different from the \(none_{1}\) and \(none_{2}\) baselines than that of cutout. However, for both cutout and cutmix, the wattle of the chicken is much more salient than the rest of the body compared to the baselines, supporting the conclusion of \cite{shah2022modeldiff} that augmentations affect the relative importance of features as well as the view of data augmentation as feature manipulation established by \cite{shen2022data}.

In order to more thoroughly study the meaningfulness of these differences, we could train pairs of networks with augmentation and certain layers of interest frozen in only one of models. If the model with the frozen weights performs more poorly than the model trained with the plastic weights, such would suggest that augmentations are critical for those layers to generate strong representations.

Furthermore, the difference in the patterns on benefits from the augmentations between ResNets trained from scratch versus fine-tuned must be explained. Given that the first layers of CNNs are known to learn low, pixel-level features that are more universal than the deeper layers, the early layers of ResNet-18 and ResNet-50 likely already learned strong, universal representations when loaded with the pre-trained ImageNet-1K weights. However, the layers near the end of the network still must learn high-level features of the new dataset, resulting in more support from the augmentations.

Thus, the role of image augmentations likely differs when used for fine-tuning versus training from a random initialization. Using image transformations for fine-tuning tasks are likely only useful for learning dataset-specific tasks. Because the studied residual networks trained from scratch received noticeable benefit from the augmentations at early layers, such transformations likely help the network learn more general image features that can be applied to a wide variety of tasks.

\section{Limitations}

\begin{figure}
    \includegraphics[width=.19\textwidth]{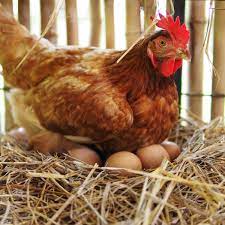}\hfill
    \includegraphics[width=.19\textwidth]{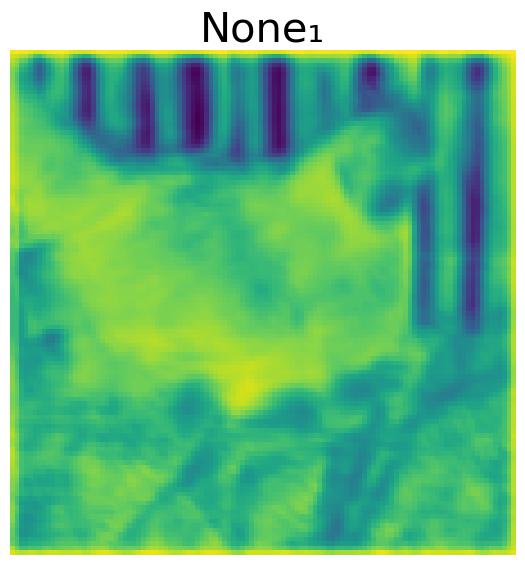}\hfill
    \includegraphics[width=.19\textwidth]{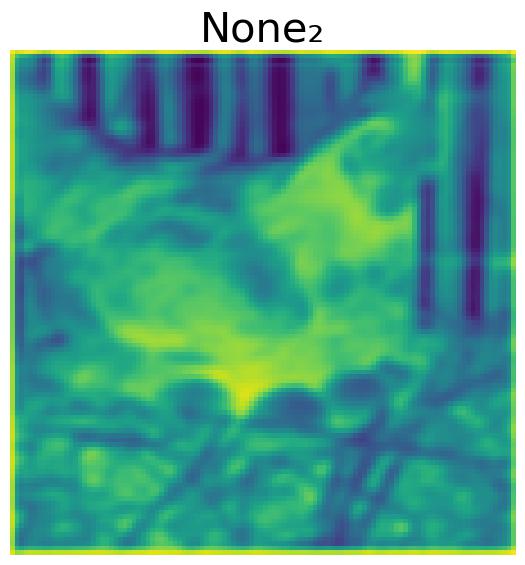}\hfill
    \includegraphics[width=.19\textwidth]{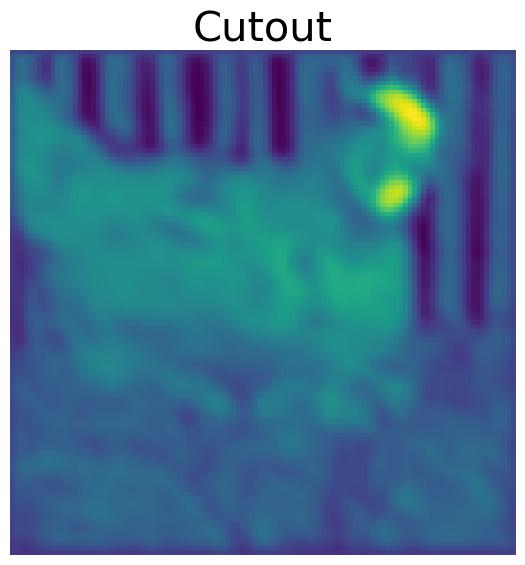}\hfill
    \includegraphics[width=.19\textwidth]{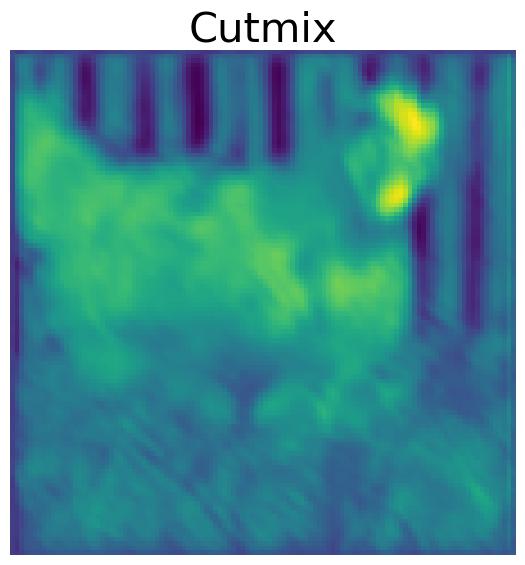}
    \caption{The 112 by 112 feature maps for a hen image produced by the 20th convolutional layer of two ResNet-56 models trained with no augmentations and two more ResNet-56s trained with cutmix and cutout.}
\end{figure}

One limitation of our study is its scope. We only study CNNs with residual connections and do not include vision transformers. Also, we use two datasets: CIFAR-10 and Tiny-Imagenet-200. While we can conclude that the width of a network and how it's initialized make significant changes to the pattern of how augmentations affect the learned weights when trained on these particular datasets, for practical purposes, more datasets should be tested, as well as real-world data. These limitations are partly due to limited computing resoures, as we tained our models and computed the CKA scores with a single NVIDIA GeForce RTX 4090 GPU on a workstation with 64GB of RAM and an Intel Core i9 CPU.

Further, we limit our study to ten augmentations on the image space with a single set of parameters for each dataset. More complex augmentations based on deep learning, such as feature space augmentations an neural style transfer, are not included within our scope.

Lastly, we use only one similarity method for comparing representations: centered kernel alignment. Although CKA does overcome limitations in comparing neural networks faced by metrics invariant to invertible linear transformation, such as canonical correlation analysis (CCA) \cite{kornblith2019similarity}, other robust coparison metrics exist, such as distance correlation \cite{zhen2022versatile}.

\section{Conclusion}

Residual neural networks learn differently from data augmentation depending on their number of layers, and some augmentations affect networks considerably more than others. When training a convolutional neural network, it may be useful to study which layers of network are most affected. For example, when some of the layers have fixed weights, as used in continual learning algorithms such as REMIND \cite{https://doi.org/10.48550/arxiv.1910.02509} and SIESTA \cite{harun2023siesta}, the weights receiving the most benefit should ideally not be kept fixed. The effectiveness of image augmentations still largely remains a black box, and it is important to study how augmentations work in addition to the performance gains they yield.

\section{Acknowledgements}

This work relied on \cite{Idelbayev18a} for their PyTorch implementation of the less-widely-known ResNets 32 and 56 as well as their code for effectively training them. Furthermore, we thank \cite{subramanian2021torch_cka} for their code to compute the CKA indices between layers of neural networks. Lastly, we thank \cite{devries2017cutout}, \cite{zhang2018mixup}, and \cite{yun2019cutmix} for their implementation of cutout, mixup, and cutmix, respectively.

{
\small
\bibliographystyle{ieee_fullname}
\bibliography{bibliography}

\begin{thebibliography}{10}\itemsep=-1pt

\bibitem{cubuk2019autoaugment}
Ekin~D. Cubuk, Barret Zoph, Dandelion Mane, Vijay Vasudevan, and Quoc~V. Le.
\newblock Autoaugment: Learning augmentation policies from data, 2019.

\bibitem{devries2017cutout}
Terrance DeVries and Graham~W Taylor.
\newblock Improved regularization of convolutional neural networks with cutout.
\newblock {\em arXiv preprint arXiv:1708.04552}, 2017.

\bibitem{elgendi2021effectiveness}
Mohamed Elgendi, Muhammad~Umer Nasir, Qunfeng Tang, David Smith, John-Paul
  Grenier, Catherine Batte, Bradley Spieler, William~Donald Leslie, Carlo
  Menon, Richard~Ribbon Fletcher, et~al.
\newblock The effectiveness of image augmentation in deep learning networks for
  detecting covid-19: A geometric transformation perspective.
\newblock {\em Frontiers in Medicine}, 8:629134, 2021.

\bibitem{harun2023siesta}
Md~Yousuf Harun, Jhair Gallardo, Tyler~L. Hayes, Ronald Kemker, and Christopher
  Kanan.
\newblock Siesta: Efficient online continual learning with sleep, 2023.

\bibitem{https://doi.org/10.48550/arxiv.1910.02509}
Tyler~L. Hayes, Kushal Kafle, Robik Shrestha, Manoj Acharya, and Christopher
  Kanan.
\newblock Remind your neural network to prevent catastrophic forgetting, 2019.

\bibitem{he2015deep}
Kaiming He, Xiangyu Zhang, Shaoqing Ren, and Jian Sun.
\newblock Deep residual learning for image recognition, 2015.

\bibitem{zhang2018mixup}
Yann N. Dauphin David Lopez-Paz Hongyi~Zhang, Moustapha~Cisse.
\newblock mixup: Beyond empirical risk minimization.
\newblock {\em International Conference on Learning Representations}, 2018.

\bibitem{Idelbayev18a}
Yerlan Idelbayev.
\newblock Proper {ResNet} implementation for {CIFAR10/CIFAR100} in {PyTorch}.
\newblock \url{https://github.com/akamaster/pytorch_resnet_cifar10}.
\newblock Accessed: 20xx-xx-xx.

\bibitem{10.1007/s10462-021-10066-4}
Nour~Eldeen Khalifa, Mohamed Loey, and Seyedali Mirjalili.
\newblock A comprehensive survey of recent trends in deep learning for digital
  images augmentation.
\newblock {\em Artif. Intell. Rev.}, 55(3):2351–2377, mar 2022.

\bibitem{kornblith2019similarity}
Simon Kornblith, Mohammad Norouzi, Honglak Lee, and Geoffrey Hinton.
\newblock Similarity of neural network representations revisited, 2019.

\bibitem{8388338}
Agnieszka Mikołajczyk and Michał Grochowski.
\newblock Data augmentation for improving deep learning in image classification
  problem.
\newblock In {\em 2018 International Interdisciplinary PhD Workshop (IIPhDW)},
  pages 117--122, 2018.

\bibitem{nguyen2021wide}
Thao Nguyen, Maithra Raghu, and Simon Kornblith.
\newblock Do wide and deep networks learn the same things? uncovering how
  neural network representations vary with width and depth, 2021.

\bibitem{PANWAR2020109944}
Harsh Panwar, P.K. Gupta, Mohammad~Khubeb Siddiqui, Ruben Morales-Menendez, and
  Vaishnavi Singh.
\newblock Application of deep learning for fast detection of covid-19 in x-rays
  using ncovnet.
\newblock {\em Chaos, Solitons \& Fractals}, 138:109944, 2020.

\bibitem{7426413}
Sérgio Pereira, Adriano Pinto, Victor Alves, and Carlos~A. Silva.
\newblock Brain tumor segmentation using convolutional neural networks in mri
  images.
\newblock {\em IEEE Transactions on Medical Imaging}, 35(5):1240--1251, 2016.

\bibitem{10.1007/978-3-319-24574-4_28}
Olaf Ronneberger, Philipp Fischer, and Thomas Brox.
\newblock U-net: Convolutional networks for biomedical image segmentation.
\newblock In Nassir Navab, Joachim Hornegger, William~M. Wells, and
  Alejandro~F. Frangi, editors, {\em Medical Image Computing and
  Computer-Assisted Intervention -- MICCAI 2015}, pages 234--241, Cham, 2015.
  Springer International Publishing.

\bibitem{santurkar2020breeds}
Shibani Santurkar, Dimitris Tsipras, and Aleksander Madry.
\newblock Breeds: Benchmarks for subpopulation shift, 2020.

\bibitem{shah2022modeldiff}
Harshay Shah, Sung~Min Park, Andrew Ilyas, and Aleksander Madry.
\newblock Modeldiff: A framework for comparing learning algorithms, 2022.

\bibitem{shen2022data}
Ruoqi Shen, Sébastien Bubeck, and Suriya Gunasekar.
\newblock Data augmentation as feature manipulation, 2022.

\bibitem{article}
Connor Shorten and Taghi Khoshgoftaar.
\newblock A survey on image data augmentation for deep learning.
\newblock {\em Journal of Big Data}, 6, 07 2019.

\bibitem{subramanian2021torch_cka}
Anand Subramanian.
\newblock torch\_cka, 2021.

\bibitem{tang2022explaining}
Jerry Tang, Manasi Sharma, and Ruohan Zhang.
\newblock Explaining the effect of data augmentation on image classification
  tasks, 2022.

\bibitem{yun2019cutmix}
Sangdoo Yun, Dongyoon Han, Seong~Joon Oh, Sanghyuk Chun, Junsuk Choe, and
  Youngjoon Yoo.
\newblock Cutmix: Regularization strategy to train strong classifiers with
  localizable features, 2019.

\bibitem{zhen2022versatile}
Xingjian Zhen, Zihang Meng, Rudrasis Chakraborty, and Vikas Singh.
\newblock On the versatile uses of partial distance correlation in deep
  learning, 2022.

\end{thebibliography}
}

\end{document}